\documentclass[rn,11pt]{ucl_document} 
\usepackage{graphicx}
\usepackage{hyperref} 

\begin{document}

\title{The State and Future of Genetic~Improvement}

\author{W. B. Langdon, et al.}

\date{25 June 2019}
\documentnumber{19/02}

\maketitle

\begin{abstract}
\vspace*{-2ex}
We report the discussion session at the sixth international
Genetic Improvement workshop, GI-2019 @ ICSE,
which was held as part of the
$41^{\rm st}$ ACM/IEEE International Conference on Software
Engineering
on Tuesday 28$^{\rm th}$ May 2019.
Topics included GI representations,
the maintainability of evolved code,
automated software testing,
future areas of GI research,
such as co-evolution,
and existing GI tools and benchmarks.
\vspace*{-2cm}
\end{abstract}

\section{Introduction}

The sixth Genetic Improvement workshop (GI 2019) was held 
as part of the 
forty-first International Conference on Software Engineering
(ICSE 2019)
in Montreal on Tuesday $28^{\rm th}$ May 2019.
Although there is a formal proceedings~\cite{Petke:2019:ICSEworkshop} for
peer-reviewed research, the workshop also included significant discussions
of current and future topics and challenges relevant to researchers and
practitioners in genetic improvement. Because the workshop included 
participants representing industrial and academic interests, our hope is
that the ideas and patterns identified may serve as one angle on the
current state of the art and 
this technical report will
serve as recorded evidence that certain
problems and topics are important. 

The next section reports the discussion at the GI 2019
workshop.
Section~\ref{sec:Freeware} lists tools that might be useful to
genetic improvement researchers.
Finally in the 
Appendix~\ref{sec:giweb} 
Nicolas Chausseau
suggests (post event)
ways for actively continuing the GI discussion on line
and suggests more biological inspirations.

\section{Discussion}

\subsection{Underexplored Research Techniques --- Evolutionary Computing} 

There is a need to investigate other forms of genetic operators,
i.e.\ other forms of mutation and crossover,
and indeed other forms of search.
such as Tabu search
and the use of anti-patterns to direct search away from unproductive
areas
\cite{Tan:2016:ASP:2950290.2950295}.

There is a need to measure what is happening inside
our GI populations
and to study the course of evolutionary search within them.

How can GI explain what it has done?
As an aside perhaps ``explain'' is being taken seriously.
Perhaps we should not start aiming at a gold-plated Rolls-Royce
``explanation.''
It appears to be adequate in some legal circumstances,
rather than quoting details of statutes or common law,
to simply say we did X in your case
because previously in a similar case to yours we did X.

There was a discussion about taking more inspiration from Biology.
Others felt simple search (e.g.\ random search) could be good
enough,
particularly if we used localisation tools to focus the search.
E.g.\ focus just on buggy code.
Concern about epistatic interactions between parts of code,
crossover etc.
might be more appropriate when considering full program synthesis.
There was also concern for the lack of Biological knowledge
amongst software engineers and the wide and ever growing
range of junk search techniques 
\cite{Weyland:2010:RAH:2433391.2433395}
which claimed Nature as their sole
justification.

\subsection{Underexplored Research Techniques --- General} 

To what extent can we extract beneficial non-AST information
from source code (or other sources),
such as comments
and 
variable naming conventions.
How can this second information channel
assist code or test transplantation,
code repair or improvement?

Perhaps there is a need to investigate if GI can
maintain code modified by GI\@.
This provoked a discussion of how to define
``Alien'' code, 
i.e.\
code generated by machines or AI.
It was suggested alien code might be defined by using
a group of software engineers 
to review the code and numerically score its comprehensibility,
if the code received a low overall score then it is alien.

Not all code is long-lived
and there may be a role for GI to generate small bits of short-lived
ephemeral code.
In particular to rapidly automatically generate a patch to a system to
keep it going
(cf.\ resilient system).
This might be required for remote systems,
or simply to keep a system running over a weekend
until it could be ``properly'' manually repaired.
Such {\em opaque} patches
need not be comprehensible,
whereas 
patches that are expected to persist in the codebase
alongside code produced by (human) developers, 
are known as {\em transparent} patches
\cite{Timperley:2018:GI}.

\pagebreak
To what extent can a co-evolutionary virtuous circle, such as starting 
with a system as its test suite, be established?
Could mutation testing~\cite{howden82}
highlight deficiencies in the test suite,
followed by 
evolution (perhaps based on EvoSuite~\cite{fraser:evosuite})
to generate new tests to extend the test suite
and so cover the now exposed gaps in testing.
Extended testing might expose bugs
which automatic program repair might generate patches for.
Is this one of our 6~impossible things for GI to do next?

\subsection{Underexplored Application Areas} 

There has already been some interest in using GI to reduced energy
consumption.
It appears some feel that this is likely only to give
small gains,
whereas others were more enthusiastic.

Can 
\href{https://en.wikipedia.org/wiki/A/B_testing}
{A/B testing} be incorporated into GI?

Others felt that software engineering should concentrate upon software
maintenance.
Potentially evolving source code
helps with software maintenance.
For example,
the first GIed code in use (BarraCUDA~\cite{Langdon:2016:GPEM})
contains evolved C code
which had been accepted by the owners of BarraCUDA
and is now maintained in the usual way.
The same is also true of
RNAfold~\cite{Langdon:2017:GI,langdon:2019:EuroGP}
and two bugfixing systems \cite{Haraldsson:thesis,Alshahwan:2019:GI}
where GI code is being maintained by people using the same tools
that they use to maintain human written code.

There was discussion about the need to maintain 
programs written in COBOL,
for which human skills are in short supply.

\subsection{Tool Selection and Integration}

How to get the best when working with other tools?
Can tools, like Daikon~\cite{ernst:dynamically-tse} and other program analysis tools, 
help the GI search
or do they simply add too much noise?

Facebook intend to make open source its automatic bugfixing tool
SapFix~\cite{Marginean:2019:ICSE},
perhaps in mid~2020.

It was suggested that compiler writers are not comfortable with
incorporating machine learning into their compilers,
particularly if code patches are incomprehensible.
However others felt this was overblown and,
for example, the University of Edinburgh compiler research group
have actively considered heuristics and learning as part of ways to
generate better 
optimised (typically meaning faster) machine code.
Indeed
genetic programming has been proposed
to give better register allocation~\cite{StephensonAMO03}
and machine learning for 
branch prediction~\cite{Muchnick:1998:ACD:286076}
and 
for Java garbage collection~\cite{andreasson2002collect}.

\subsection{Benchmark Selection and Generality} 

Is there a danger of working on programs that are too small?
Are small benchmark program atypical?
Are they more fragile~\cite{langdon:2015:csdc}
than programs that people care about?

Although GI has been demonstrated with
assembler, byte code and binary machine code,
most GI work has operated on the level of source code.
The two favoured approaches have been to use the 
abstract syntax tree~(AST)
generated by the compiler
and representing the source code via a grammar.
What other ways of representing the genetic material
should be considered?
E.g.\
Jhe-Yu Liou et~al.~\cite{Liou:2019:GI}
suggested using
compiler intermediate representation~(IR),
as generated by the LLVM compiler.
Whilst
control flow graphs~\cite{Muchnick:1998:ACD:286076}
might also be adopted by GI as an evolvable representation
\cite{Stadler2013GraalI}.

It was felt important to make results freely available.
The workshop web pages 
\href{http://geneticimprovementofsoftware.com/faq/}
{http://geneticimprovement\allowbreak{}of\allowbreak{}software.com/faq/}
already points to some open source tools and benchmarks.
Web sites such as GitHub
and tools like Docker containers
might help and also reduce the impact of 
``\href{https://en.wikipedia.org/wiki/Software_rot}{bit rot}''
on older tools.

\pagebreak
There was  discussion of the importance of benchmarks.
Some felt it made life easier,
e.g.\  to reproduce others results
and therefore they aided good science.
Others, however, felt that the point of evolutionary search was to try
(and hopefully succeed)
in solving problems which had previously not been considered at all or
where simply regarded as being impossible.
A counter argument was put that we can never get enough
tests to define precisely what the program being tested will do in all
circumstances.
In other words the semantics of the program have to be defined.
Can we treat testing as sampling the programs behaviour and
in someway say how we expect the program to behave between samples?
I.e.,
how should it interpolate or
extrapolate
from test cases?

\subsection{Industrial Interest} 

Employment prospects.
Student and post-doctoral internships are available at Facebook
and elsewhere

In some programs an approximate answer
may be sufficient, i.e.\ good enough.

\newpage
\section{Tools and Resources}
\label{sec:Freeware}

Some free tools that were mentioned during 
the GI-2019 workshop
(extended by the co-authors in the three weeks after GI-2019).

\begin{enumerate} 

\item
\href{http://geneticimprovementofsoftware.com/faq/}
{http://geneticimprovementofsoftware.com/faq/}

\item
\href{https://en.wikipedia.org/wiki/Genetic_improvement_(computer_science)}
{Wikipedia Genetic improvement (computer science)}

\item
\href{https://en.wikipedia.org/wiki/Automatic_bug_fixing}
{Wikipedia Automatic bug fixing}

\item
Google VirusTotal 
\href{http://www.virustotal.com}
{www.virustotal.com}

\item
\label{i.llvm}
LLVM, 
especially LLVM's C compiler intermediate representation
\href{https://github.com/eschulte/llvm-mutate}
{https://github.com/eschulte/\allowbreak{}llvm-mutate}

\item
Rodinia benchmarks
\href{https://rodinia.cs.virginia.edu}
{https://rodinia.cs.virginia.edu}

\item
ThunderSVM machine learning
\href{https://github.com/Xtra-Computing/thundersvm}
{https://github.com/Xtra-Computing/thundersvm}

\item
Sleuth2 pattern mining
\href{https://cran.r-project.org/web/packages/Sleuth2}
{https://cran.r-project.org/web/packages/Sleuth2}

\item
GraalVM
\href{https://www.graalvm.org/}
{https://www.graalvm.org/}
and
truffle (truffle~3)

\item
Defects4J
\href{https://github.com/rjust/defects4j}
{https://github.com/rjust/defects4j}

\item
Tiny CC, tcc may give faster compilation.
I.e.\ trade compilation time against execution time.
Similarly we may not want all compiler optimisation switched on
during GI evolution.

\item
EvoSuite 
\href{http://www.evosuite.org}
{http://www.evosuite.org}

\item
ManyBugs and IntroClass,
Benchmark of small buggy programs for~C
\href{https://repairbenchmarks.cs.umass.edu/}
{https://repairbenchmarks.\allowbreak{}cs.\allowbreak{}umass.edu/}

\item
IntroClassJava,
Benchmark of small buggy programs for Java
\href{https://github.com/Spirals-Team/IntroClassJava}
{https://github.com/Spirals-Team/\allowbreak{}IntroClassJava}

\item
GenProg
(Automatic repair tool for C)
\href{https://github.com/squaresLab/genprog-code}
{https://github.com/squaresLab/genprog-code}

\item
GenProg4Java,
(Automatic repair tool for Java)
\href{https://github.com/squaresLab/genprog4java}
{https://github.com/squaresLab/genprog4java}

\item
BugZoo is a container-based platform for studying historical bugs
\href{https://github.com/squaresLab/BugZoo}
{https://github.com/squaresLab/\allowbreak{}BugZoo}
\cite{Timperley:2018:ICSE}

\item
Darjeeling is a language-agnostic tool for search-based program repair
\href{https://github.com/squaresLab/Darjeeling}
{https://github.com/\allowbreak{}squaresLab/\allowbreak{}Darjeeling}

\item
GIN
(lightweight micro-framework for the Genetic Improvement of Java code)
\href{https://github.com/gintool}
{https://github.com/\allowbreak{}gintool}
\cite{White:2017:GI}

\item
PyGGI
(Python General Framework for Genetic Improvement)
\href{https://github.com/coinse/pyggi}
{https://github.com/coinse/pyggi}
\cite{An2017aa}

\item
SEL, Software Evolution Library
(which works with llvm-mutate, item~\ref{i.llvm} above)
\href{https://github.com/grammatech/sel}
{https://github.com/\allowbreak{}gramma\allowbreak{}tech/\allowbreak{}sel}

\item
Blue
(simple blue example of grammar based Genetic Improvement)
\href{http://www.cs.ucl.ac.uk/staff/W.Langdon/gggp/#code}
     {http://www.cs.ucl.ac.uk/staff/\allowbreak{}W.Langdon/\allowbreak{}gggp/\#code}
(see Free Code heading) 
\cite{langdon:RN1806}

\end{enumerate} 

\appendix
\section{Post Workshop Suggestions by Nicolas Chausseau}
\label{sec:giweb}

Turn this article into an ongoing collaborative website online
it could: 
\begin{itemize}
\item
    Allow ongoing discussion in a more informal online forum setting
\item
    Help discover people working on same or similar problems, 
\item
    And potentially enable collaboration, or branching out of other
    fruitful, more specific conversations on those topics. 
\item
    ... In other words it could be the starting point for new
  research, to be discussed, collaborators found, etc. 
  Even paper drafts reviewed and commented on. 
  (Nicolas Chausseau volunteered to help set up such a website).

\vspace*{1ex}
  GI might follow ML4SE's example and
  set up a Google forum.
  He wondered if others interested in GI would
  participate in such a forum (e.g.\ ask questions, make comments).
\end{itemize}

\subsection{Biological input to GP}

Turn this article into an ongoing collaborative website online
it could: 
\begin{itemize}

\item
I would be curious to know which recent papers discuss approaches
inspired from biology / nature, if any
(since there are no references for this section)

In particular I wanted to read more about these very specific topics:
Hierarchical structure of DNA (e.g. junk DNA, which we know now,
enables or disables vast libraries of SNPs, and creates epistatic
interactions) -- what are the implications for GAs? Is it fruitful at
all to store genetic information in hierarchical form?

\item
Hierarchical structure of layers of neural networks (e.g. bottom
layers evolve to become multipurpose feature detectors, during
backpropagation) -- we know in the case of neural networks this
hierarchical structure is what allowed them to solve hard problems, by
having lower layers learn these "multipurpose features".

\item
Hierarchical structure of human codebases (object-oriented code, where
a first layer of reusable building block is constantly combined and
recombined into new applications) -- can GAs use these "pre-evolved
APIs" more efficiently, to speed up convergence? Can we also store a
program's "DNA information" in a more hierarchical form somehow? Or is
human representation already optimal for building-block reuse, for
enabling and disabling pre-evolved building blocks?

\item
... It seems all 3 solutions (DNA, deep NNs and human codebases)
present a hierarchical structure, a structure that allows for
"building-block discovery" and "building-block reuse".

\begin{itemize}
\item
  I found only one article so far that examined this question a bit, it
  made interesting observations in the "Modularity" section
  https://www.hindawi.com/journals/jaea/2010/568375/\#B30, from 2010
\item
  Is it possible that the way information is stored (as a hierarchy)
  affects the speed of convergence of a population? Has this been
  examined before? Did we observe the formation of building blocks in
  GAs, has this been measured?
\end{itemize}

\item
Concerning search and GA hybrids, any additional link is welcome! [5]
is very interesting.

\end{itemize}

\bibliographystyle{splncs03}

\bibliography{gp-bibliography,references,slice}

\end{document}